\documentclass{article}
\pdfoutput=1
\usepackage{cite}
\usepackage{cprotect}
\usepackage{listings}
\usepackage{tikz}
\usepackage{ctable}

\makeatletter
\renewenvironment{table}%
{\renewcommand\familydefault\sfdefault
\@float{table}}
{\end@float}
\makeatother

\usepackage{bchart}
\usepackage{multirow}

\usepackage{placeins}
\usepackage[width=.90\textwidth]{caption}
\usepackage{geometry}
\usepackage{setspace}
\usepackage{makecell}

\usepackage{hyperref}
\hypersetup{
colorlinks=true,
linkcolor=black,
filecolor=magenta,
urlcolor=black,
citecolor=black
}

\title{\textbf{Transfer Learning from Transformers to Fake News Challenge Stance Detection (\hbox{FNC-1}) Task}}
\date{}
\author{Valeriya Slovikovskaya\\ University of Pisa\\ https://www.unipi.it}
\begin{document}
    \maketitle
    \thispagestyle{empty}

    \begin{abstract}
        In this paper, we report improved results of the Fake News Challenge Stage 1 (\hbox{FNC-1}) stance detection task.
        This gain in performance is due to the generalization power of large language models based on Transformer
        architecture, invented, trained and publicly released over the last two years.
        Specifically (1) we improved the \hbox{FNC-1} best performing model adding BERT sentence embedding of input sequences
        as a model feature, (2) we fine-tuned BERT, XLNet, and RoBERTa transformers on \hbox{FNC-1} extended dataset and
        obtained state-of-the-art results on \hbox{FNC-1} task.
    \end{abstract}

    \section{Introduction}\label{sec:introduction}
    Two years ago, the Fake News Challenge Stage 1 (\hbox{FNC-1}) attracted attention of several members of the linguistics
    research community:  fifty teams participated and submitted their results, some of them like~\cite{riedel2017fnc}
    released their code and published research papers.
    Later,~\cite{hanselowski-etal-2018-retrospective} retrospectively examined the top submissions\footnote{including
    that of Hanselowski team Athene which ranked second}, challenge task formulation, dataset characteristics as well as
    baseline model and evaluation metric proposed by organizers.
    Their contribution included revised FNC score, extended dataset and novel classification model that outperforms the
    winner's system on class-wise F1 scores.
    Besides, they also determined upper bound for \hbox{FNC-1} data classification conducting an annotation experiment
    that involved five human raters, who manually labelled 200 data instances.
    This detailed analysis of the task, together with the code released by the authors, are highly valuable as a
    baseline on which we can rely in order to obtain further improvements.

    In the last two years since Fake News Challenge Cup, significant improvements have occurred in NLP technology, in
    particular with the development of large language models using contextualized word embeddings based on the Google
    Transformer architecture~\cite{DBLP:journals/corr/VaswaniSPUJGKP17}.

    Applying transfer learning techniques to these general purpose pre-trained models helped to achieve improvements in
    a wide range of downstream tasks.
    Using these models as pre-training for downstream tasks allows lowering the amount
    of annotated data required for such tasks.

    BERT~\cite{DBLP:journals/corr/abs-1810-04805}, GPT~\cite{radford2019language},
    XLNet~\cite{DBLP:journals/corr/abs-1906-08237}, trained on the huge unlabeled datasets on GPUs and Cloud TPUs,
    attained significant accuracy
    increase on GLUE~\cite{DBLP:journals/corr/abs-1804-07461} and RACE~\cite{lai-etal-2017-race}
    benchmarks\footnote{GLUE leaderboard: https://gluebenchmark.com/leaderboard/,
    RACE leaderboard: \url{http://www.qizhexie.com/data/RACE_leaderboard.html}}.

    The public releases of pre-trained models led to intensified experimenting with transfer learning / fine-tuning for
    a broader range of linguistic tasks;
    this stimulated the development of APIs that unify and standardize the access to
    weights of different pre-trained models and made them accessible for the greatly extended community of professionals.

    This paper is organized as follows.
    In the first part, we present Fake New Challenge Stage 1 (\hbox{FNC-1}) revised task: task formulation, dataset,
    evaluation metric, and the state-of-the-art model.
    In the second part, we introduce new semantic features that allow to improve the model performance.
    In the third part we present the further
    improved results obtained as a fruit of transfer learning from BERT, RoBERTa and XLNet transformers applied to the
    \hbox{FNC-1} classification task.

    \section{\hbox{FNC-1} task}\label{sec:fake-news-challenge-stage-1-task}

    \subsection{Task formulation}\label{subsec:task-formulation}

    Let's suppose we have a statement or a claim which truthfulness is verified or which source is trusted.
    How can we then detect liars about such claim in an upcoming news stream?
    We might try to determine whether the news text is related to the claim, and, if so, to find out whether it
    contradicts or supports it, i.e.\ its stance on the subject.
    The Fake News Challenge Stage 1 task is formulated in fact as a stance detection problem.
    The data set provided by organizers consists of pairs of a news article headline (as a claim),
    and a snippet of text taken either from the same or from another news article.
    Each pair in the training part of the data set is labeled with a tag attesting the relation between the headline
    and the snippet.
    The labels are "agree", "disagree", "discuss", and "unrelated".
    Table~\ref{table:FNC1Example} presents a few examples of stance classification, taken from the training set of the
    challenge.
    The goal of the challenge is to classify the pairs in the test data set.

    \begin{table}[ht!]
        \captionsetup{font=small}
        \small
        \centering
        \begin{tabular}{p{14mm} p{98mm}}
            Headline  & Robert Plant Ripped up \$800M Led Zeppelin Reunion Contract    \\
            \specialrule{.12em}{.05em}{.05em}
            agree      & Led Zeppelin's Robert Plant turned down \pounds500 MILLION to reform supergroup  \\  \hline
            disagree   & No, Robert Plant did not rip up an \$800 million deal to get Led Zeppelin back together   \\
            \hline
            discuss    & Robert Plant reportedly tore up an \$800 million Led Zeppelin reunion deal    \\  \hline
            unrelated  & Richard Branson's Virgin Galactic is set to launch SpaceShipTwo today  \\ \hline
        \end{tabular}
        \captionsetup{justification=centering}
        \caption{Headline and text snippets from document bodies with respective stances from the \hbox{FNC-1} dataset}
        \label{table:FNC1Example}
    \end{table}

    \subsection{Dataset}\label{subsec:dataset}

    \subsubsection{\hbox{FNC-1} data}

    The organizers of the Challenge constructed a news corpus with data from the Emergent dataset for stance classification
    that contains 300 claims, and 2,595 associated article headlines.

    The claims were collected by journalists from a variety of sources such as rumour sites, e.g. Snopes.com, and Twitter
    accounts such as @Hoaxalizer.
    Their subjects include world and national U.S.\ news and technology stories.
    The journalist also summarises the article into a headline.
    In parallel to the article-level stance detection, a
    claim-level veracity judgement is reached as more articles associated with the claim are examined.

    The dataset construction was conducted by Craig Silverman and his colleagues at the Tow Center for Digital Journalism
    at Columbia University\footnote{http://towcenter.org/} for Emergent Project~\cite{CraigSilverman}.
    Then it was used
    for independent research on rumor debunking and released\footnote{https://github.com/willferreira/mscproject} as
    Emergent dataset by~\cite{Ferreira2016EmergentAN}

    Thus the Emergent Project dataset consists of 300 topics, each represented by a claim with 5--20 possibly related
    news articles.
    The labelling of a headline/article pair indicates whether the article is supporting, refuting, or
    just reporting the claim, respectively.

    \begin{table}[ht!]
        \captionsetup{font=small}
        \small
        \centering
        \begin{tabular}{ l r r r r r r r }
            Dataset & headlines & documents & instances & agree & disagree & discuss & unrelated \\
            \hline
            \hbox{FNC-1}   & 2,587     & 2,587     & 75,385    & 7.4\% & 2.0\%    & 17.7\%  & 72.8\%    \\
        \end{tabular}
        \captionsetup{justification=centering}
        \caption{Corpus statistics and label distribution for the \hbox{FNC-1} dataset}
        \label{table:FNC1Corpus}
    \end{table}

    Other than for rumor debunking, the \hbox{FNC-1} organizers match each document with every summarized headline, and then
    label the pair with one of the four stance labels: "agree", "disagree", "discuss", and "unrelated".
    To generate the "unrelated"
    class, headlines and articles belonging to different topics are randomly matched.
    Headline/article pairs of 200
    claims/topics are reserved for training, the remaining headline/article pairs of 100 claims/topics for testing.
    Claims, headlines, and articles are therefore not shared between the two data splits.
    To prevent teams from using
    any unfair means by deriving the labels for the test set from the publicly available Emergent data, the organizers
    additionally created 266 instances.
    Table~\ref{table:FNC1Corpus} shows the corpus size and label distribution.

    \subsubsection{ARC data}

    To test the robustness of their models (i.e. how well they generalize to new datasets),
   ~\cite{hanselowski-etal-2018-retrospective} introduced a novel test data for document-level stance
    detection based on
    the Argument Reasoning Comprehension (ARC) task proposed by~\cite{DBLP:journals/corr/abs-1708-01425}.

   ~\cite{DBLP:journals/corr/abs-1708-01425} manually selected 188 debate topics with popular questions from
    the user debate section of the New York Times.
    For each topic, they collected user posts, which are highly
    ranked by other users, and created two claims representing two opposing views on the topic.
    Then, they asked crowd
    workers to decide whether a user post supports either of the two opposing claims or does not express a stance at all.
    This Argument Reasoning Comprehension (ARC) dataset consists of typical controversial topics from the news domain,
    such as immigration, schooling issues, or international affairs.

    \begin{table}[ht!]
        \captionsetup{font=small}
        \small
        \centering
        \begin{tabular}{ | p{8mm} | p{50mm} | p{50mm} | }
            \hline
            \multicolumn{3}{|c|}{\textbf{Example from the original ARC dataset}} \\ \hline
            Topic     & \multicolumn{2}{|p{100mm}|}{Do same-sex colleges play an important role in education, or are
            they outdated?} \\ \hline
            User post & \multicolumn{2}{|p{100mm}|}{Only 40 women’s colleges are left in the U.S. \ And, while there are
            a variety of opinions on their value, to the women who have attended \dots them, they have been \dots tremendously
            valuable. \dots} \\ \hline
            Claims    & \multicolumn{2}{|p{100mm}|}{1. Same-sex colleges are outdated 2.
            Same-sex colleges are
            still relevant} \\ \hline
            Label     & \multicolumn{2}{|p{100mm}|}{Same-sex colleges are still relevant} \\ \hline
            \multicolumn{3}{|c|}{\textbf{Generated instance in alignment with the \hbox{FNC-1} problem setting}} \\ \hline
            Stance & Headline                             & Document \\ \hline
            agree  & Same-sex colleges are still relevant & Only 40 women's colleges are left in the U.S. \dots \\ \hline
        \end{tabular}
        \captionsetup{justification=centering}
        \caption{Example of the original ARC dataset and the generated instance to align with \hbox{FNC-1} dataset}
        \label{table:ARCExample}
    \end{table}

    While this is similar to the \hbox{FNC-1} dataset, there are significant differences, as a user post is typically a
    multi-sentence statement representing one viewpoint on the topic.
    In contrast, the news articles of \hbox{FNC-1}
    are longer and usually provide more balanced and detailed perspective on the issue.

    To allow using the ARC data for the \hbox{FNC-1} stance detection setup~\cite{hanselowski-etal-2018-retrospective}, consider each
    user post as an article and randomly select one of the two claims as the headline.
    They label the headline/article pair
    as "agree" if the claim has also been chosen by the workers, as "disagree" if the workers chose the opposite claim,
    and as "discuss" if the workers selected neither claim.
    Table~\ref{table:ARCExample} shows an example of the
    revised ARC corpus structure.

    In order to generate the unrelated instances~\cite{hanselowski-etal-2018-retrospective}, randomly match the user
    posts with claims, but avoid that a user post is assigned to a claim from the same topic.
    For training and testing,
    the authors split the corpus into 80\% training/validation set and 20\% testing set.
    Table~\ref{table:ARCCorpus} provides basic corpus statistics.

    \begin{table}[ht!]
        \captionsetup{font=small}
        \small
        \centering
        \begin{tabular}{ l r r r r r r r }
            Dataset & headlines & documents & instances & agree & disagree & discuss & unrelated \\ \hline
            ARC & 4,448     & 4,448     & 17,792    & 8.9\% & 10.0\%   & 6.1\%   & 75.0\%    \\
        \end{tabular}
        \captionsetup{justification=centering}
        \caption{Corpus statistics and label distribution for the ARC dataset}
        \label{table:ARCCorpus}
    \end{table}

    For all our experiments we use the combined FNC + ARC dataset made freely available
    by~\cite{AndreasHanselowskiGitHub} on GitHub.

    \subsection{Evaluation metric}\label{subsec:evaluation-metric}

    The \hbox{FNC-1} dataset is highly unbalanced with respect to class distribution.
    The \hbox{FNC-1} organizers proposed the hierarchical evaluation metric, FNC score, which first awards .25 points if an
    article is correctly classified as related or "unrelated" to a given headline.
    If it is related, .75 additional points are assigned if the model correctly classifies
    the headline/article pair as "agree", "disagree", or "discuss".
    The goal of this weighting schema is to balance out the large number of unrelated instances.

    This metric was criticized by~\cite{hanselowski-etal-2018-retrospective} as the score that, albeit an hierarchical,
    fails to take into account the imbalanced class distribution of the three related classes "agree", "disagree", and
    "discuss".

   ~\cite{hanselowski-etal-2018-retrospective} reasoned that the models which perform well on the majority class
    and poorly on the minority classes are favored.
    Since it is not difficult to separate related from unrelated instances (the best systems reach about F1 = .99 for the
    "unrelated" class), a classifier that just randomly predicts one of the three related classes would already achieve a
    high FNC score.
    A classifier that always predicts "discuss" for the related documents even reaches FNC = .833, which is
    even higher than the top-ranked FNC system.

    Authors therefore argued that the FNC score metric is not appropriate for the task and proposed to rely on the
    class-wise and the macro-averaged F1 scores (F1m) not affected by the large size of the majority class.
    The na\"{\i}ve approach of perfectly classifying "unrelated" and always predicting "disagree" for the related classes
    would achieve only F1m = .444 and would reveal a poor ability of the model to distinguish between "agree",
    "disagree", and "discuss" categories.

    \subsection{Model architecture}\label{subsec:model-architecture}

    With respect to F1m score, the model of team Athene~\cite{HanselowskiPVSSchillerCaspelherr} outperforms all other
    models.
    Its architecture, named "featMLP", is
    a multi-layer perceptron (MLP) inspired by the work of~\cite{RichardDavisAndChrisProctor}.
    The model has six
    hidden and a softmax layers and incorporates multiple hand-engineered features.
    The one group of features includes unigrams, and the cosine similarity of word embeddings of nouns and verbs between
    headline and document tokens.
    The other group consists of topic models based on non-negative matrix factorization, latent Dirichlet allocation,
    and latent semantic indexing. featMLP also incorporates the \hbox{FNC-1} baseline's features suggested by the
    \hbox{FNC-1} organizers such as word overlap, polarity words, refuting words, and co-occurrence feature that counts
    how many times word 1-/2-/4-grams, character 2-/4-/8-/16-grams, and stop words of the headline appear in the first
    100, first 255 characters of the article, and how often they appear in the article overall.
    See~\cite{hanselowski-etal-2018-retrospective} and baseline~\cite{ByronGalbraithAndHumzaIqbal} for major details.
    Depending on the feature type, they either form separate feature vectors for document and headline, or a joint feature
    vector.

    \subsection{Lexical features}\label{subsec:lexical-features}

    Authors conduct the feature ablation test on the \hbox{FNC-1} development set with 10-fold cross-validation to define the
    best feature set for their model.
    This feature set contains bag-of-word (BoW) and bag-of-character (BoC) features as well as topic modeling features.

    For BoW features authors use uni- and bi-grams with 5,000 tokens vocabulary for the headline as well as for the article.
    Based on a technique by~\cite{RePEc:inm:ormnsc:v:53:y:2007:i:9:p:1375-1388} they add a negation tag "\_NEG" as prefix
    to every word between special negation keywords (e.g. "not", "never", "no") until the next punctuation mark appears.
    For the BoC, three-grams are chosen with 5,000 tokens vocabulary, too.
    For the BoW/BoC feature authors use the TF
    to extract the vocabulary and to build the feature vectors of headline and document. The resulting TF vectors of
    headline and article get concatenated afterwards.
    In the set of winning Bow/BoC features, authors also include \hbox{FNC-1} baseline's co-occurrence feature.

    As for topic models, authors use non-negative matrix factorization (NMF)~\cite{ChihJenLin}, latent semantic indexing
    (LSI)~\cite{Deerwester90indexingby}, and latent Dirichlet allocation (LDA)~\cite{Blei:2003:LDA:944919.944937}
    to create topic models out
    of which they create independent features.
    For each topic model, they extract 300 topics out of the headline and
    article texts.
    Afterwards, they compute the similarity of headlines and article bodies to the found topics separately and
    either concatenate the feature vectors (NMF, LSI) or calculate the cosine distance between them as a single valued
    feature (NMF, LDA).

    Based on these features the model exploits similarity between the headline and the article in terms of rather
    lexical overlap without really capturing the semantics of the text.

    \subsection{Semantic features}\label{subsec:semantic-features}

    Stepping towards taking into account semantic characteristics of the text,
   ~\cite{hanselowski-etal-2018-retrospective} experiment with "stackLSTM" model, which combines the best feature set
    found in the ablation test with a stacked long short-term memory (LSTM) network~\cite{NIPS2013_5166}.

    Authors use 50-dimensional GloVe word
    embeddings\footnote{\url{http://nlp.stanford.edu/data/glove.twitter.27B.zip}}~\cite{pennington2014glove} in
    order to generate sequences of word vectors of a headline/article pair.
    For this, they concatenate a maximum of 100 tokens of the headline and the article.
    These embedded word sequences are fed through
    two stacked LSTMs with a hidden state size of 100 with a dropout of 0.2 each.
    The last hidden state of the second
    LSTM is concatenated with the feature set and fed into a 3-layer neural network with 600 neurons each.
    Finally, they add a dense layer with four neurons and softmax activation function in order to retrieve the class
    probabilities.

    "stackLSTM" improved performance for the "disagree" class, which is the most difficult one to predict due to the
    low number of instances.

    \section{\hbox{FNC-1} top model improved}\label{sec:fnc-1-top-model-improved}

    In our work on improving \hbox{FNC-1} results we were looking for semantically sensitive model architectures and features.
    Therefore, we experimented with transfer learning from large pre-trained models reported as being able to capture
    semantics of the text.

    There are two existing strategies for applying pre-trained language representations to downstream tasks:
    feature-based and fine-tuning.
     The feature-based approach uses task-specific architectures that include the
    pre-trained representations as additional features.

    The fine-tuning does not require task specific model, instead it introduces into a general purpose pre-trained
    model a restricted set of task specific parameters and trains it on the downstream task dataset, in general
    for 2--3 epochs, fine-tuning all pre-trained parameters.

    We apply both approaches to \hbox{FNC-1} task, and report them both to be effective.

    \subsection{InferSent based features}\label{subsec:infersent-based-features}

   ~\cite{conneau-EtAl:2017:EMNLP2017} indicated the suitability of natural language inference (NLI) for transfer
    learning to other NLP tasks: NLI is a high-level understanding task that involves reasoning about the semantic
    relationships within sentences.
    Authors trained universal sentence representations using the supervised data of the
    Stanford Natural Language Inference (SNLI) dataset~\cite{bowman-etal-2015-large} and made their encoder, based
    on a bi-directional LSTM architecture with max pooling, publicly available so that sentence embedding can be easily
    obtained and transferred as a feature to other text classification tasks.

    Thus, for each headline/article body pair of FNC-ARC dataset, we calculated vector representation for every sentence
    of the article or headline using Facebook InferSent encoder~\cite{AlexisConneauGitHub}, then we averaged sentence
    vectors to obtain vector representation for whole article and whole headline, after that we concatenated the article
    and the headline vectors.
    We used the headline/article pair vectors as a feature on its own to directly
    feed into~\cite{hanselowski-etal-2018-retrospective} "featMLP" classifier and in addition, as a separate
    features, we calculated a cosine similarity score between headline and article vectors as well as a maximum
    similarity score between headline and every sentence in the article.

    These three features together slightly improve F1 macro and F1 -"agree", -"disagree" and -"discuss" scores of
    "featMLP" model as shown in Table~\ref{table:FNCmodelimproved}.
    "Inf1" column reports the impact of InferSent
    embeddings for input sequence pair, "Inf3" shows the impact of this feature together
    with two similarity scores for two sequences in the input pair.

    \begin{table}[ht!]
        \captionsetup{font=small}
        \small
        \centering
        \begin{tabular}{  l  r  r  r  r  r  r | r }
            Score, \% & Base & Inf1 & Inf3 & BERT1 & BERT3 & \begin{tabular}{@{}c@{}}BERT \\+Inf3\end{tabular} & \begin{tabular}{@{}c@{}}BERT \\ only\end{tabular} \\
            \hline
            Accuracy     & 87.18  & 86.93   &  87.05     & 87.57          & 87.75   & \textbf{87.94}  &   81.45  \\
            FNC score    & 78.59  & 77.82   &  78.16     & 79.03          & 79.42   & \textbf{79.72}  &  68.65  \\
            \hline
            F1 macro     & 61.61  & 61.63   &  62.02     & 63.32 & \textbf{63.96}   &          63.91  &  56.27  \\
            F1 agree     & 50.26  & 50.66   &  50.69     & 53.77 &         53.15    &          53.58  &  43.16  \\
            F1 disagree  & 27.22  & 27.81   &  28.85     & 31.43 & \textbf{34.04}   &          31.22  &  31.75  \\
            F1 discuss   & 72.43  & 71.70   &  72.20     & 71.21 &         71.61    &  \textbf{73.88} &  58.55  \\
            F1 unrelated & 96.54  & 96.35   &  96.36     & 96.89 & \textbf{97.05}   &          96.95  &  91.61  \\
        \end{tabular}
        \captionsetup{justification=centering}
        \caption{Performance of featMLP model without and with InferSent- and BERT- based features}
        \label{table:FNCmodelimproved}
    \end{table}

    \subsection{BERT based features}\label{subsec:bert-based-features}

    Our next step was experimenting with transfer learning from pre-trained transformers.

    \cite{DBLP:journals/corr/abs-1810-04805} introduced a new language representation model called BERT, which
    stands for \textbf{B}idirectional \textbf{E}ncoder \textbf{R}epresentations from \textbf{T}ransformers.
    BERT's model
    architecture is a multi-layer bidirectional Transformer encoder based on the original implementation described in
    ~\cite{DBLP:journals/corr/VaswaniSPUJGKP17} and released in the tensor2tensor library~\cite{tensor2tensor}.
    BERT is designed to pre-train deep bidirectional representations from unlabeled text by jointly conditioning on both
    left and right context in all layers.
    BERT uses "masked language model" (MLM) pre-training objective, inspired by
    the Cloze task~\cite{doi:10.1177/107769905303000401}.
    Masked language model randomly masks
    some of the tokens from the input, and the objective is to predict the masked word based on its context.
    In addition to the masked language model, BERT also use a "next sentence prediction" (NSP) task that jointly
    pre-trains text-pair representations.

    For the pre-training corpus~\cite{DBLP:journals/corr/abs-1810-04805} use the BooksCorpus (800M words)
   ~\cite{DBLP:journals/corr/ZhuKZSUTF15} and English Wikipedia
    (2,500M words).\footnote{Authors note that it is critical to use a document-level corpus
    rather than a shuffled sentence-level corpus such as the Billion Word Benchmark
   ~\cite{DBLP:journals/corr/ChelbaMSGBK13} in order to extract long contiguous sequences.}

    Self attention mechanism in the Transformer allows BERT to model many downstream tasks -- whether they involve single
    text or text pairs.
    For each task, the steps are: (1) simply plug in the task specific inputs and outputs into BERT
    and (2) fine-tune all the parameters end-to-end.

    Authors showed that fine-tuned BERT outperforms all systems on all GLUE tasks by a substantial
    margin: BERT-base obtains a 4.5\% average accuracy improvement over the prior state of the art, and BERT-large
    achieves 7.0\%.

    \cite{DBLP:journals/corr/abs-1810-04805} report fine-tuning to be relatively inexpensive compared to pre-training:
    at most 1 hour on a single Cloud TPU, or a few hours on a GPU, starting from the same pre-trained model. Authors
    pre-train BERT model of two sizes: BERT-base with 12 transformer blocks, 12 attention heads, 768 neurons in hidden
    layers, and 110 million parameters, and BERT-large with 24 transformer blocks, 16 attention heads, 1024 neuron in
    hidden layers, and 340 million parameters.
    Both models are made publicly available and their success raised
    the popularity of libraries like HuggingFace's Transformers~\cite{Wolf2019HuggingFacesTS} that facilitate access
    to pre-trained models and optimize their integration into NLP pipelines.

    In our experiments, we follow a feature-based approach first to transfer learning and use BERT sentence embeddings
    to create the new features for "featMLP" model.
    As with InferSent sentence embeddings, we introduce three separate features: we calculate BERT sentence embeddings for
    each sentence of an article and a headline, then we average sentence embeddings to obtain vector representation
    for the whole article and the whole headline, after that we concatenate the article and the headline vectors.
    We use the headline/article pair vectors as a feature on its own, and, in addition, as two separate features, we
    calculate the cosine similarity score between a headline and article vectors, and the maximum
    similarity score between headline and each sentence of the article.

    To obtain BERT sentence embeddings we leverage the "bert-as-service" application
   ~\cite{xiao2018bertservice}\footnote{with default settings listed here:
    \url{https://github.com/hanxiao/bert-as-service}}.
    "bert-as-service" is build on pre-trained 12/24-layer BERT models released by Google AI, it uses BERT as a sentence
    encoder and hosts it as a service via ZeroMQ. The application includes build-in HTTP server and a dashboard;
    it provides asynchronous encoding and multicasting.

    The performance of the "featMLP" model enriched with BERT-based features is shown in Table~\ref{table:FNCmodelimproved}.
    "BERT1" column shows the impact of BERT vector representation of an input, and "BERT3" shows the impact of this feature
    together with two similarity scores between BERT embeddings of two sequences in the input pair.
    The column "BERT3 + Inf3" reports InferSent and BERT cumulative effect.
    The most significant F1-score increase, that of 7\% is attained for most difficult "disagree" class.

    To demonstrate the power of BERT embeddings we conduct a kind of ablation test: we remove all Bow/BoC
    and topic modelling features and use only headline/article
    pair BERT vectors (without similarity scores), this way we obtaine the results shown in the "BERT1 only" column of
    Table~\ref{table:FNCmodelimproved}: it can be seen that the "disagree" F1 score improvement is relied mostly
    on the BERT embedding feature.

    These results pushed us further on the way of experimenting with pre-trained transformers.
    Following~\cite{DBLP:journals/corr/abs-1810-04805} we fine-tuned on \hbox{FNC-1} task three pre-trained models: BERT,
    XLNet and RoBERTa,  and as expected, received substantially improved F1 scores with respect to
   ~\cite{hanselowski-etal-2018-retrospective} "featMLP" model performance.

    \section{Transfer learning: from transformers to \hbox{FNC-1} task}\label{sec:transfer-learning:-from-transformers-to-fnc-1-task}

    \subsection{XLNet}\label{subsec:xlnet}

    BERT predicts all masked positions independently, meaning that during the training, it does not learn to handle
    dependencies between predicted masked tokens.
    This reduces the number of dependencies BERT learns at once, making
    the learning signal weaker than it could be.

    Another problem with BERT is that the [MASK] token -- which is at the center of training BERT -- never appears when
    fine-tuning BERT on downstream tasks.
    That means that the [MASK] token is a source of train-test skew while fine-tuning.

    XLNet~\cite{DBLP:journals/corr/abs-1906-08237} incorporates bidirectional context while avoiding the [MASK]
    tokens and independent predictions.
    It does this by introducing "permutation language modeling": instead of
    predicting the tokens in sequential order, it predicts tokens in some random order.\cprotect\footnote{For example, the
    sequence order is $x_1$, $x_2$, $x_3$, $x_4$.
    So for this 4 (N) tokens  sentence, there are 24 (N!) permutations.
    If we want to predict the $x_3$, so there are 4 patterns with $x_3$ in the 24 permutations:
    $x3$ is at the 1st position, 2nd position, 3rd position, and 4th position: [$x_3$, xx, xx, xx],
    [xx, x3, xx, xx], [xx, xx, x3, xx], [xx, xx, xx, x3].
    Here we set the position of $x_3$ as t-th position and t-1
    tokens are the context words for predicting $x_3$.
    Intuitively, the model will learn to gather information from all positions on both sides.}

    Aside from using permutation language modeling, XLNet improves upon BERT by using the Transformer XL as its base
    architecture.

    Both BERT and XLNet can take the pair of text sequences as an input.
    To enable the model to distinguish between
    words in two different segments, BERT learns a segment embedding.
    In contrast, XLNet learns an embedding that represents whether two words are from the same segment.
    This embedding is used during attention computation between any two words.

    XLNet integrates the segment recurrence mechanism and relative encoding scheme of TransformerXL
   ~\cite{DBLP:journals/corr/abs-1901-02860} into pre-training and enables the model to reuse hidden states from
    previous segments.\footnote{XLNet is named after TransformerXL}

    XLNet outperforms BERT and achieves state-of-the-art performance across 20 tasks including text classification,
    question answering, natural language inference, duplicate sentence (question) detection, document ranking, coreference
    resolution.

    \subsection{RoBERTa}\label{subsec:roberta}

    \cite{DBLP:journals/corr/abs-1907-11692} found that BERT was significantly undertrained and proposed an improved
    recipe for training BERT models, which they call RoBERTa (Robustly optimized BERT approach), that can match or
    exceed the performance of all of the post-BERT methods.
    The recipe includes: (1) training the model longer, with
    bigger batches, over more data; (2) removing the "next sentence prediction" objective; (3) training on longer
    sequences;
    and (4) dynamically changing the masking pattern applied to the training data.

    To train RoBERTa authors use five English-language corpora of varying sizes and domains, totaling over
    160GB of uncompressed text.\footnote{These corpora include (1) BOOK CORPUS~\cite{DBLP:journals/corr/ZhuKZSUTF15}
    plus English Wikipedia,
    the original data used to train BERT (16GB); (2) CC-NEWS, which authors collected from the English portion of the
    CommonCrawl News dataset~\cite{SebastianNagel}, the data contains 63 million English news articles crawled
    between September 2016 and February 2019 (76GB after filtering); (3) OPEN WEB TEXT~\cite{Gokaslan2019OpenWeb},
    an open-source recreation of the WebText corpus described in~\cite{radford2019language}, the text is web content
    extracted from URLs shared on Reddit with at least three upvotes (38GB); (4) STORIES, a dataset introduced
    in~\cite{DBLP:journals/corr/abs-1806-02847} containing a subset of CommonCrawl data filtered to match the
    story-like style of Winograd schemas (31GB).}

    They also demonstrate that removing the "next sentence prediction" loss together with segment-pair input format
    matches or slightly improves downstream task performance.
    RoBERTa achieves state-of-the-art results on all 9 of the
    GLUE task development sets.
    It consistently outperforms both BERT and XLNet.

    \subsection{Transformers fine-tuned on \hbox{FNC-1} task}\label{subsec:transformers-fine-tuned-on-fnc-1-task}

    For our experiments on fine-tuning transformers on \hbox{FNC-1} task we used the Simple Transformers
   ~\cite{ThilinaRajapakse} wrapper around Hugging Face Transformers library~\cite{Wolf2019HuggingFacesTS}.

    \begin{table}[ht!]
        \captionsetup{font=small}
        \small
        \centering
        \begin{tabular}{  l  r  r  r  r }
            Score, \%     & featMLP & BERT   &  XLNet  & RoBERTa  \\
            \hline
            Accuracy      & 87.18   & 91.32  &  92.10  & 93.19    \\
            FNC score     & 78.59   & 86.16  &  87.90  & 89.17    \\
            \hline
            F1 macro      & 61.61   & 72.82  &  76.06  & 78.12    \\
            F1 agree      & 50.26   & 64.70  &  68.61  & 70.71    \\
            F1 disagree   & 27.22   & 47.84  &  54.85  & 58.06    \\
            F1 discuss    & 72.43   & 80.07  &  82.10  & 84.57    \\
            F1 unrelated  & 96.54   & 98.68  &  98.65  & 99.16    \\
        \end{tabular}
        \captionsetup{justification=centering}
        \caption{Performance of featMLP, BERT, XLNet, and RoBERTa fine-tuned models}
        \label{table:FineTunedTransformersF1scores}
    \end{table}

    The development of Transformers, a library for natural language processing with transfer learning models, was motivated by
    the need to share state-of-the-art pre-trained models and to increase the availability of published research code.
    With this library low-resource users can reuse pre-trained models without having to train them from scratch.

    \begin{table}[ht!]
        \footnotesize
        \captionsetup{font=small}
        \centering
        \begin{tabular}{  |l|r|r|r|r| }
            \hline
            & agree   & disagree  &  discuss  & unrelated \\
            \hline
            agree   & \textbf{1104 1534 1597 1592} & 152 328 297 265 & 738 619 474 377 & 243 24 50 32   \\
            \hline
            disagree  &  306 151 114 129 & \textbf{230 432 517 560}  &  253 136 150 150  &  280 18 35 21   \\
            \hline
            discuss  &  617 486 482 487 & 175 239 205 210  & \textbf{3337 3732 3894 4009}  &  514 222 262 132  \\
            \hline
            unrelated &  129 66 44 29 &  64 70 50 34  &  244 156 125 107  &  \textbf{20586 20759 20676 20838} \\
            \hline
        \end{tabular}
        \captionsetup{justification=centering}
        \caption{Confusion matrix: every cell shows featMLP, BERT, XLNet, RoBERTa result correspondingly}
        \label{table:FineTunedTransformersConfusionMatrix}
    \end{table}

    These models are accessed through a simple and unified API that follows a classic NLP pipeline: setting up
    configuration, processing data with a tokenizer and encoder, and using a model either for training (adaptation in
    particular) or inference.
    The model implementations provided in the library are tested to ensure they match the
    original author implementations' performances on various benchmarks.
    A list of architectures for which reference implementations and pre-trained weights are currently provided in
    Transformers includes BERT~\cite{DBLP:journals/corr/abs-1810-04805}, DistilBERT~\cite{sanh2019distilbert},
    RoBERTa~\cite{DBLP:journals/corr/abs-1907-11692}, XLNet~\cite{DBLP:journals/corr/abs-1906-08237},
    GPT and GPT2~\cite{radford2019language}.

    \begin{table}[ht!]
        \small
        \captionsetup{font=small}
        \centering
        \begin{tabular}{ lrrrr }
            &   Precision, \%  &    Recall, \%  &  F1-score, \%  &  Support \\
            \hline
            0    &   \textbf{53} 69 71 71   &  \textbf{48} 61 66 70  &  \textbf{50} 65 69 71  &    2237  2505  2418  2266 \\
            1    &   \textbf{38} 40 48 52   &  \textbf{24} 59 63 65  &  \textbf{29} 48 55 58  &    1069   737   816   860 \\
            2    &   \textbf{73} 80 84 86   &  \textbf{74} 80 80 83  &  \textbf{73} 80 82 85  &    4643  4679  4843  4838 \\
            3    &   \textbf{95} 99 98 99   &  \textbf{98} 99 99 99  &  \textbf{97} 99 99 99  &   21023 21051 20895 21008 \\
        \end{tabular}
        \captionsetup{justification=centering}
        \caption{Precision, recall, F1-score for featMLP, BERT, XLNet, RoBERTa. Every cell contains 4 results
        corresponding to four models in order}
        \label{table:FineTunedTransformersPrecisionRecallF1Score}
    \end{table}

    The Simple Transformers~\cite{ThilinaRajapakse} library is built on top of the Hugging Face Transformers.
    The idea behind it was to make it as simple as possible, abstracting a lot of the implementation details.

    Thus, with Simple Transformers on the shoulders of Hugging Face Transformers we could access pre-traines BERT, XLNet
    and RoBERTa in unified way without a lot of pre-processing coding.

    We modified only limited number of parameters, setting maximum sequence length to be equal to 512 tokens
    \footnote{The maximum possible value to set given the parameters of pre-trained models}.
    We used the base versions of
    all transformers ("bert-base-uncased", "xlnet-base-cased", "roberta-base"), those with 12 transformer blocks.
    We trained every transformers for 5 epochs with batch size of 4, with learning rate of 3e-5 for BERT and 1e-5 for
    XLNet and RoBERTa.\footnote{Every time fine-tuning was taking several hours on Nvidia Tesla P100 GPU with 16GB of
    memory.}

    Our results, reported in Tables~\ref{table:FineTunedTransformersF1scores},
    ~\ref{table:FineTunedTransformersConfusionMatrix}, and~\ref{table:FineTunedTransformersPrecisionRecallF1Score}
    show that all three models fine-tuned on the
    Fake News Challenge task outperform~\cite{hanselowski-etal-2018-retrospective} "featMLP" model with significant margins
    varying from 8 to 20\% for "related" classes.
    The XLNet and RoBERTa demonstrate better performance than BERT,
    that corresponds to our expectations;
    RoBERTa outperforms both BERT and XLNet, that is in line with the results
    reported in~\cite{DBLP:journals/corr/abs-1907-11692}.

    We also performed cross domain validation experiments fine-tuning transformers on \hbox{FNC-1} dataset and
    evaluating them on ARC dataset and vice versa.
    The Table~\ref{table:FNCARC_ARCFNC} shows that for these test RoBERTa can demonstrate worse performance in
    comparison with BERT and XLNet models.

    \begin{table}[ht!]
        \captionsetup{font=small}
        \small
        \centering
        \begin{tabular}{ l  r  r  r  r }
            & \multicolumn{4}{l}{FNC - ARC}\\
            F1 score, \% & featMLP & BERT & XLNet & RoBERTa \\
            \specialrule{.12em}{.05em}{.05em}
            agree      & 25.71  & 39.55 & 43.82 & 45.61  \\
            disagree   &  9.67  & 28.39 & 28.31 & 24.91 \\
            discuss    & 15.42  & 18.58 & 08.30 & 18.23 \\
            unrelated  & 89.83  & 93.48 & 92.53 & 93.91 \\
        \end{tabular}
        \begin{tabular}{  l  r  r  r  r }
            & \multicolumn{4}{l}{ARC - FNC}\\
            F1 score, \% & featMLP & BERT & XLNet & RoBERTa \\
            \specialrule{.12em}{.05em}{.05em}
            agree      & 27.57  &  39.03  & 41.01  &   0.73  \\
            disagree   & 10.43  &   8.41  & 05.44  &  16.10  \\
            discuss    &  8.03  &   5.66  & 14.07  &  10.40  \\
            unrelated  & 85.43  &  95.72  & 95.59  &  93.57  \\
        \end{tabular}
        \captionsetup{justification=centering}
        \caption{Class-wise F1 scores for "featMLP" model (with BoW/BoC and topic modeling features) and fine-tuned
        transformers in cross-domain, FNC - ARC and ARC - FNC, experiments}
        \label{table:FNCARC_ARCFNC}
    \end{table}

    \section{Conclusion}\label{sec:conclusion}

    We substantially improved Fake News Challenge Stage 1 results, putting the task in the line of Natural Language
    Processing tasks that are reported to benefit from transfer learning from pre-trained transformers.

    \section{Acknowledgments}\label{sec:acknowledgments}

    The author would like to thank Professor Giuseppe Attardi, who provided insight for this work, and Professor Serge
    Sharoff and Dr. Pavel Braslavsky for their helpful advise.

    The experiments were carried out on a server with 4 GPUs Nvidia Tesla Pascal P100, partially funded by the
    University of Pisa through the grant Grandi Attrezzature 2016.

    \bibliography{article}{}
    \bibliographystyle{apalike}
\end{document}